# Benchmarking Differential Evolution on a Quantum Simulator

Parthasarathy Srinivasan a,1

[a] Oracle Corporation USA
Orchid ID: Parthasarathy Srinivasan
https://orcid.org/0000-0002-2352-5649

**Abstract.** The use of Evolutionary Algorithms (EA) for solving Mathematical/Computational Optimization Problems is inspired by the biological processes of Evolution. Few of the primitives involved in the Evolutionary process/paradigm are selection of 'Fit' individuals (from a population sample) for retention, cloning, mutation, discarding, breeding, crossover etc. In the Evolutionary Algorithm abstraction, the individuals are deemed to be solution candidates to an Optimization problem and additional solution(/sets) are built by applying analogies to the above primitives (cloning, mutation etc.) by means of evaluating a 'Fitness' function/criterion. One such algorithm is **Differential Evolution(DE)** which can be used to compute the minima of functions such as the rastrigin function and rosenbrock function. This work is an attempt to study the result of applying the **DE** method on these functions with candidate individuals generated on classical Turing modeled computation and comparing the same with those on state of the art Quantum computation.The study benchmarks the convergence of these functions by varying the parameters initialized and reports timing, convergence, and resource utilization results.

1. ## Introduction

Differential Evolution is a Genetic Algorithm technique to solve Combinatorial Optimization problems, by mutating random candidate solutions (population) and evolving the solution pool over several generations. The fitness function considered in this work is the restrain and rosenbrock functions which are mathematically represented respectively as follows:

$f(X) = \text{Sum}[x_j^2 - 10*\cos(2*pi*x_j)] + 10n$   **(Formula 1(rastrigin))**

$f(X) = \sum i=1-(n-1)((100(x_{i+1} - x_i)^2)^2 + (x_i - 1)^2), -30 \le x_i \le 30$   **(Formula 2 (rosenbrock))**

In this work the best solution to the above function is found by evolving a pool of random candidate solutions, by applying mutation (picking 3 random candidate solutions viz. a,b,c and computing the mutant with the following computation:

mutation[k] = population[a][k] + F *(population[b][k] - population[c][k])   **(Formula 3)**

(k varying from 1 to 3 and F being a constant)

The solution pool is then evolved by computing the rastrigin error and thresholding the mutant.

---

1 Corresponding Author : Parthasarathy Srinivasan , parthasarathy.s.srinivasan@oracle.com

## 2. Scope of Work

The present work is pre-requisite for further research in the field of evolutionary algorithm implementation on Quantum machines involving Human Interaction because to study fitness functions with unknown form (which is essentially HCI), it is first necessary to fully understand known fitness function behavior on Quantum Systems such as the rastrigin function and rosenbrock function which is preliminarily investigated in this work.

## 3. Motivation and significance of this work

The Evolutionary Algorithm methodology/paradigm has been known and employed widely to several computational problems for the past couple of decades, enjoying reasonable success with algorithms implemented on 'Classical' (Turing machine) based computers. The work submitted herein seeks to (originally) evaluate the outcome of attempting to extend implementation of the Critical Operation of the Differential Evolution approach on 'Quantum' Computers which are based on Qubits and Quantum Registers.

From a review of existing scientific literature, gaps have been found on this theme of comparison thus motivating creation of new work as an attempt to start addressing this gap.

In this direction this work induces a frame of reference for **orthogonal and comparative analyses**, to comprehend the impact on system behavior; when there is a switch in the paradigm used for computing solutions to mathematical functions of presently known form , and potential extensions of those analyses to the problem area of Human Computer Interaction which involves computing solutions to mathematical functions whose form and behavior is presently unknown/less known.

From this perspective the present work is a new beginning to be used as a moot point for furthering the aforementioned field of scientific discussion.

This will hopefully provide the foundation for (example) applications in the field of HCI in **biomedical engineering** etc.

## 4. Definitions/Glossary (source: wiki):

**Differential Evolution**
In evolutionary computation, differential evolution (DE) is a method that optimizes a problem by iteratively trying to improve a candidate solution with regards to a given measure of quality. Such methods are commonly known as metaheuristics as they make few or no assumptions about the problem being optimized and can search exceptionally large spaces of candidate solutions. However, metaheuristics such as DE do not guarantee an optimal solution is ever found.

**Quantum Computer**

A quantum computer is a computer that exploits quantum mechanical phenomena. At small scales, physical matter exhibits properties of both particles and waves, and quantum computing leverages this behavior using specialized hardware. Classical physics cannot explain the operation of these quantum devices, and a scalable quantum computer could perform some calculations exponentially faster than any modern "classical" computer.

**Qubit (or QBit)**

A Qubit is a unit of measure used in quantum computing. Like a bit in normal (non-quantum) computing, a Qubit has two distinct states, 0 state and the 1 state. However, unlike the normal bit, a qubit can have a state that is somewhere in-between, called a "superposition." You cannot measure the superposition without the superposition going away (changing).

## 5. Contextualization and Impact

The more diverse and truer a population of random numbers is **[McCaffrey [1]],** due to coherence the more amenable its solution space is to capture and exploiting the underlying 'genetic' data pattern for solution convergence.

1 Corresponding Author : Parthasarathy Srinivasan , parthasarathy.s.srinivasan@oracle.com

This work sets the stage to perform a spot check of how this plays out in the Quantum space by providing a starting checkpoint to measurably evaluate the approach and opens up the discussions for improvements.

Adding to the same lines this work provides empirical experimental data substantiating advantage of QRNG based populations **[Microsoft [2]]** to be amenable to less generational iteration for solution.

**[Xie [8]]** Attempts to find fundamental relationships and tuning options for "Parent Selection Pressure".

On these lines **[[Xie [8]]** states :"The second is whether a selection scheme supports parallel architectures because a parallel architecture is very useful for speeding up learning paradigms that are computationally intensive."

This work dwells on the possibility of exploiting the inherent parallelism in QRNG due to the existence of a Qubit in a state of superposition rather than the classical either 1 or 0 constraint and hence influencing convergence and quality of random number generation.

To further explain the above position, we offer below Comparison of Classical and Quantum Systems w.r.t. Random number based applications

**Pros of using Quantum Systems (Comparing and complementing existing works with this new application oriented one)**

   **a.** The quality of random numbers generated by a quantum system is higher than that of a classical system w.r.t. low correlation and long periodicity. Following is an extract from the literature in **[Xiangfan-Ma [4]]**:

"Quantum physics can be exploited to generate true random numbers, which have important roles in many applications, especially in cryptography. Genuine randomness from the measurement of a quantum system reveals the inherent nature of quantumness—coherence, an important feature that differentiates quantum mechanics from classical physics."

   The above premise in the works is complemented by this work by evaluating the applicability and results using standard functions and comparing their implementations on Classical and Quantum models.
   The results thus obtained indeed bear testimony to the correctness of the hypothesis as shown by standard statistical analyses presented further.

   **b.** Quantum RNG results in quicker convergence (the number of generations required is less though the final exit criterion kicks in much later than classical machines). Following is an extract from the literature in **[Medium [5]]**:

"A wide variety of quantum computing algorithms have been proposed with the perspective of obtaining solutions to some of the classically unsolvable problems. The currently available quantum computers are far away from its utilization for commercial applications, yet many algorithms are developed for near term applications and study of optimization and speedup capability of quantum processing unit."

   The work conducted in this paper clearly supplements the need faced above and attempts to start filling the gaps.

   **c.** QRNG increases the chance of obtaining more relevant and usable solutions because of the above quality of random numbers.
   Following is an extract from the literature in **[Jacak [6]]**:

"2013—Google confirmed that the IBM Java SecureRandom class in Java Cryptography Architecture (JCA) generated repetitive (and therefore predictable) sequences, which compromised

1 Corresponding Author : Parthasarathy Srinivasan , parthasarathy.s.srinivasan@oracle.com

application security made for Android to support the electronic currency Bitcoin"

The work carried out as part of this paper opens doors to initially research and sooner more than later commercial utilization of QRNG, thus overcoming much of the teething problems to a direct and complete application of Quantum computing technology to solve real world problems.

**Pros of using Classical systems** (These are substantiated by abundant material in public domain.)

a. Inexpensive to use when the application is not demanding/critical for higher solution quality instead is tolerable such as financial engineering or purely academic/development purpose in nature.

b. When there is a huge base of not so critical legacy solutions which currently rely on classical machines and will require considerable time and cost to get migrated to quantum machines before yielding substantial ROI.

c. Useful in applications where any solution obtained is usable and there is no advantage/bias to use one solution as against the other.

**6. Differential Evolution (DE) (Algorithm) pseudocode**

Pseudocode of function invoking the rastrigin and rosenbrock functions (Fitness functions for thresholding mutant individuals(candidate solutions generated by mathematically combining parent individuals))

1. Set up the mutating and thresholding parameters:

   $dim = 3$   (Number of dimensions of the population individual's input parameter to the fitness function)
   $pop\_size = 50$ (Number of individuals (candidate solutions) in the input)
   $F = 0.5$   # mutation (The multiplier used in the recombination of individuals)
   $cr = 0.7$   # crossover (The threshold meta heuristic which determines the inclusion of the new mutant individual by replacing an existing member of the population)
   $max\_gen = 200$ (The number of generational iterations before which the best solution generated using the logic as above is accepted as final).

2. Initialize the population of candidate solution individuals by sampling the probability from the quantum simulator(by means of executing H Gate on qubits) (for the quantum case) and from the numpy library(for the classical case).

   (N.B.:- Certain nuances must be dealt for the rosenbrock function due to the constraint on its definition only between population values -30 to 30 and this influences some convergence results)

3. Compute the population error vector by calling the rastrigin or rosenbrock error computation function which returns the error difference between the functions' known global minima and the functional value for individual members of the population.

4. Randomly pick 3 population individuals and mathematically combine them to produce new mutant individual.

5. Compute the error for the new mutant individual by calling the rastrigin or rosenbrock function on the mutant.

6. Randomly generate another probability and threshold it with the crossover to accept the mutant by replacing a member of the population or rejecting the mutant.

1 Corresponding Author : Parthasarathy Srinivasan , parthasarathy.s.srinivasan@oracle.com

7. Repeat steps 4 to 6 for the number of maximum generations setup in step 1 keeping a record of the generational best solution (solution with minimum error).

8. Obtain the final best solution by finding the solution with minimum error from the generational best solutions.

**7. Convergence results.**

The results herein lend unequivocal credence to the fact that Quantum Random Number Generators generate a better quality of Random numbers which result in significantly better convergence results as shown by the summary and illustrations below:

On the **X axis** is plotted the number of maximum generational iterations of the DE Algorithm (described above)

On the **Y axis** is plotted the index of the population individual where the best solution (minimum error) has been found

It is evident that the rastrigin and rosenbrock functions converge quicker (more favorably/optimally) when QRNG is utilized as opposed to Classical (Turing) model.

**For rastrigin**

QRNG convergence point :   (6,10)

Classical convergence point : (11,17)

**For rosenbrock**

QRNG convergence point :   (11, 50)

Classical convergence point : (85,0)  [Corrected per re-observation]

**Supplemental illustrations**

**QRNG Convergence for rastrigin**

1 Corresponding Author : Parthasarathy Srinivasan , parthasarathy.s.srinivasan@oracle.com

**Classical Convergence for rastrigin**

**QRNG Convergence for rosenbrock)**

1 Corresponding Author : Parthasarathy Srinivasan , parthasarathy.s.srinivasan@oracle.com

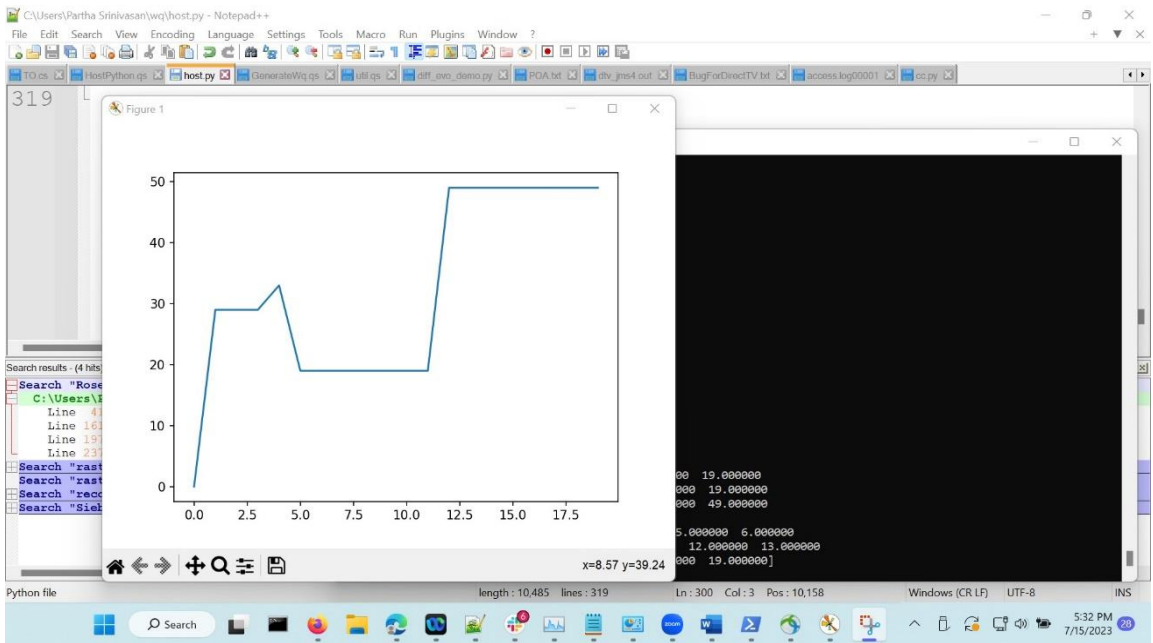

**Classical convergence for rosenbrock**

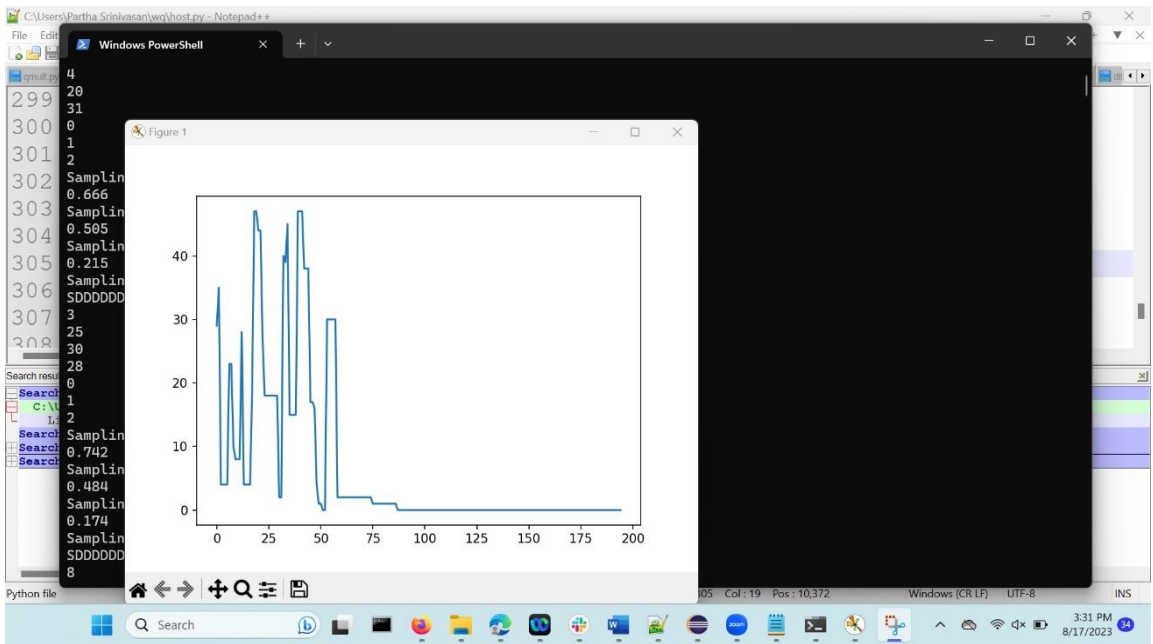

**8 a) Following is result of statistical comparison (using www.statskingdom.com) of :**

**Group 1 : Quantum Rastrigin , Group 2 : Quantum Rosen (RNG quality derived)**

1 Corresponding Author : Parthasarathy Srinivasan , parthasarathy.s.srinivasan@oracle.com

**Two sample Mann Whitney U test, using Normal distribution (two-tailed) (validation)**

The **normal approximation** is used. The statistic's distribution is $N(24.5, 7.766^2)$.
The data contains ties, identical values, it is recommended to use the **normal approximation** that uses the ties correction.

**1. H$_0$ hypothesis**
Since p-value > α, H$_0$ cannot be rejected.
The randomly selected value of **Group1's** population is assumed to be **equal to** the randomly selected value of **Group2's** population.
In other words, the difference between the randomly selected value of **Group1** and the **Group2** populations is not big enough to be statistically significant.

**2. P-value**
The p-value equals 0.07142, ( $p(x \leq Z) = 0.03571$ ). It means that the chance of type I error, rejecting a correct H$_0$, is too high: 0.07142 (7.14%).
The larger the p-value the more it supports H$_0$.

**3. The statistics**
The test statistic Z equals -1.8028, which is in the 95% region of acceptance: [-1.96 : 1.96].
U=10, is in the 95% region of acceptance: [9.2793 : 39.7207].

**4. Effect size**
The observed **standardized effect size**, $Z/\sqrt{(n_1+n_2)}$, is **medium** (0.48). That indicates that the magnitude of the difference between the value from **Group1** and the value from **Group2** is medium.
The observed **common language effect size**, $U_1/(n_1 n_2)$, is **0.2**, this is the probability that a random value from **Group1** is greater than a random value from **Group2**.

**8 b) Following is result of statistical comparison of :**

**Group 1 : Quantum Rastrigin , Group 2 : Classical Rastrigin (RNG quality derived)**

**Two sample Mann Whitney U test, using Normal distribution (two-tailed) (validation)**

The **normal approximation** is used. The statistic's distribution is $N(24.5, 7.783^2)$.
The data contains ties, identical values, it is recommended to use the **normal approximation** that uses the ties correction.

**1. H$_0$ hypothesis**
Since p-value > α, H$_0$ cannot be rejected.
The randomly selected value of **Group1's** population is assumed to be **equal to** the randomly selected value of **Group2's** population.
In other words, the difference between the randomly selected value of **Group1** and the **Group2** populations is not big enough to be statistically significant.

**2. P-value**
The p-value equals 0.1231, ( $p(x \leq Z) = 0.06156$ ). It means that the chance of type I error, rejecting a correct H$_0$, is too high: 0.1231 (12.31%).
The larger the p-value the more it supports H$_0$.

**3. The statistics**
The test statistic Z equals -1.5418, which is in the 95% region of acceptance: [-1.96 : 1.96].
U=12, is in the 95% region of acceptance: [9.2454 : 39.7546].

**4. Effect size**
The observed **standardized effect size**, $Z/\sqrt{(n_1+n_2)}$, is **medium** (0.41). That indicates that the magnitude

1 Corresponding Author : Parthasarathy Srinivasan , parthasarathy.s.srinivasan@oracle.com

of the difference between the value from **Group1** and the value from **Group2** is medium.
The observed **common language effect size**, $U_1/(n_1n_2)$, is **0.24**, this is the probability that a random value from **Group1** is greater than a random value from **Group2**.

**8 c) Discussion on statistical analysis of convergence data**

The interpretation of the above two statistical analysis clearly and consistently establishes the fact that Quantum RNG is higher in quality than classical RNG as viewed by a comparison of the p-value from the cases above : Case 1 -> p-value 0.07142 (comparison of the 2 quantum cases) ; Case 2 (comparison of 1 classical and the corresponding quantum case) -> p-value 0.1231 . This is substantiated by the derivation that in Case 1 the chance of Type I error is ~ half of that in Case 2 and consequently better convergence.

## 9. Timing results (rastrigin)

**Table 1. (Timing values of probabilistic(classical) operation in seconds)**

| Sample size 10 | Sample size 50 | Sample size 100 |
| --- | --- | --- |
| 0.0029854774475097656 | 0.00725 | 0.015338 |
| 0.00759577751159668 | 0.088405609130 | 0.013422727584838 |
| 0.013752460479736328 | 0.052 | 0.0591559410095214 |

**Table 2. (Timing values of quantum (probability generation operation in seconds). (Number of generations 10, 50,100) on y axis.**

| Sample size 10 | Sample size 50 | Sample size 100 |
| --- | --- | --- |
| 0.0029854774475097656 | 0.00725 | 0.00759577751159668 |
| 0.013752460479736328 | 0.088405609130 | 0.052 |
| 0.015338 | 0.013422727584838 | 0.0591559410095214 |

**10. Resource Utilization (As expected)**

**Classical (Turing Model)**

CPU usage : 3% at 1.17 GHz.
Memory usage : 13.1 of 44.9 GB

**Quantum (RNG)**

CPU usage : 15% at 1.07 GHz.
Memory usage : 12.9 of 44.9 GB

**Supplemental Illustrations**

**Classical (Turing Model)**

1 Corresponding Author : Parthasarathy Srinivasan , parthasarathy.s.srinivasan@oracle.com

**Quantum (RNG)**

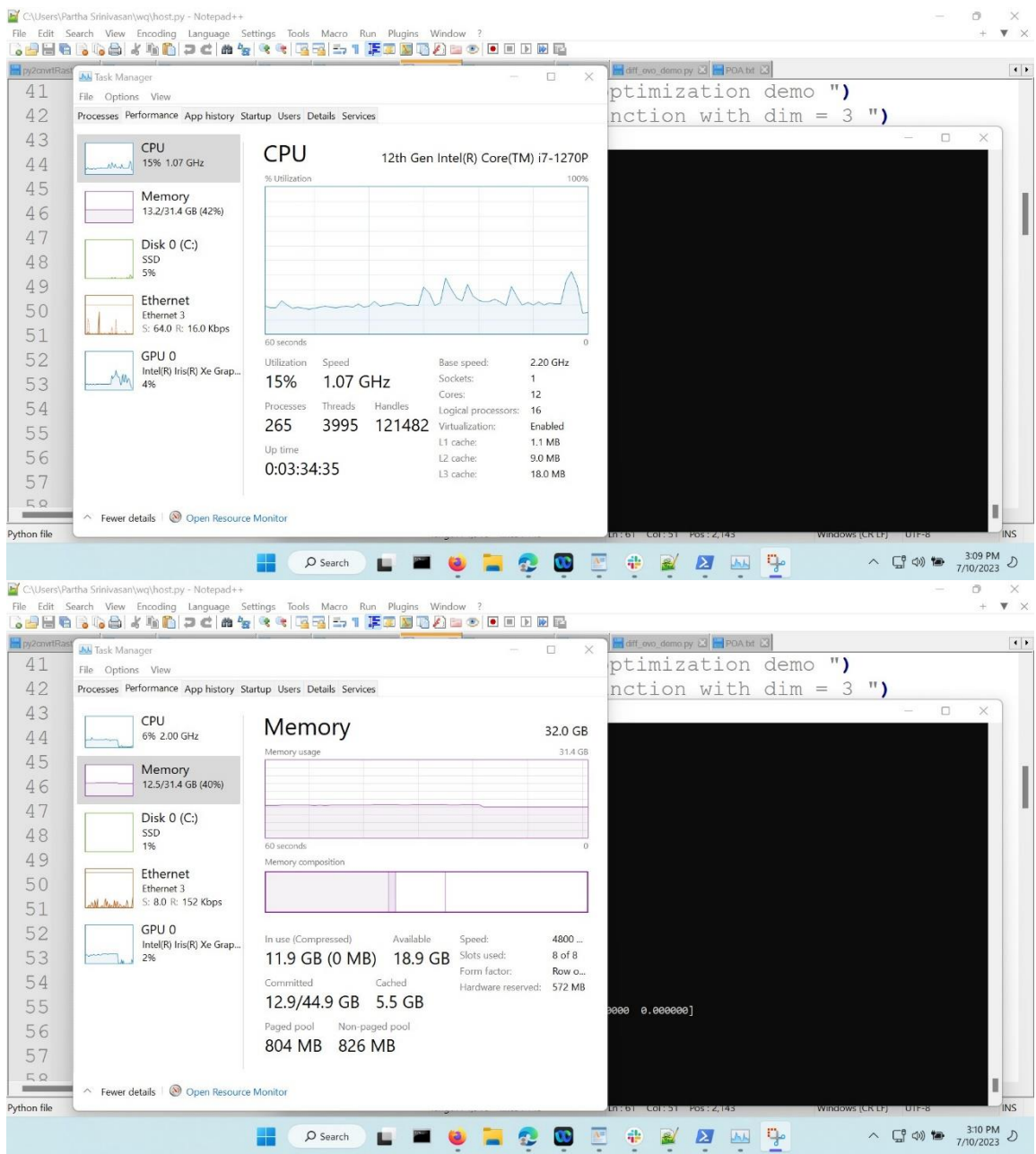

## 11. Conclusion

Contrary to intuition the quantum simulator does not live up to the timing measure when compared against the classical computer for the critical probability generation operation. Perhaps a real quantum backend may surpass this. However due to integration and availability problems the Microsoft Azure quantum backends could not be utilized in this work but may be part of derivative works. Quantum machines have a little ways to go.

Having said so the quality of random numbers produced by the simulator is evidently better and the implications such as convergence and timing have been discussed herewith in section 8. (Reproduced here for ready reference)

**Discussion on statistical analysis of convergence data**

The interpretation of the above two statistical analysis clearly and consistently establishes the fact that Quantum RNG is higher in quality than classical RNG as viewed by a comparison of the p-value from the cases above : Case 1 -> p-value 0.07142 (comparison of the 2 quantum cases) ; Case 2 (comparison of 1 classical and the corresponding quantum case) -> p-value 0.1231 . This is substantiated by the derivation that in Case 1 the chance of Type I error is ~ half of that in Case 2 and consequently better convergence.

12. QRNG (provider) Current Limitations and Scope for Future work

   **Integration Problem**:

   3 of the 4 quantum available backends have Integration Problems at present in that they are not capable of returning Integer values. This work serves to make this problem visible and public so that this gets addressed in future quantum offerings by quantum providers
   thus, enabling a smoother design and development facilitation for quantum computation consumers.

   **Availability Problem**: Many of the high end quantum backends have limited availability at present again as found through this work and those that are available don't integrate suitably as above.

   Future work should reasonably involve real quantum backend machines for obtaining results for comparison in place of the quantum simulator utilized in this work.

   Also, it is suggested use other benchmark functions (few of which are mentioned by **[Charilogis[4]]**) and cross verify results both with quantum simulators and real quantum backends.